# Cognitive Ledger Project: Towards Building Personal Digital Twins Through Cognitive Blockchain

Amir Reza Asadi
Humind Labs
Tehran,Iran
a.asadi@humind.xyz
ORCID: 0000-0001-5440-8456

*Abstract*—The Cognitive Ledger Project is an effort to develop a modular system for turning users' personal data into structured information and machine learning models based on a blockchain-based infrastructure. In this work-in-progress paper, we propose a cognitive architecture for cognitive digital twins. The suggested design embraces a cognitive blockchain (Cognitive ledger) at its core. The architecture includes several modules that turn users' activities in the digital environment into reusable knowledge objects and artificial intelligence that one day can work together to form the cognitive digital twin of users.

## I. Introduction

Digital Twin is a trending topic in manufacturing technology studies. It refers to the virtual representation of a physical object or system that simulates the behavior of the physical thing [1]. However, this paper does not deal with the manufacturing industry. In this project, we develop a cognitive ledger for use by personal cognitive digital twins of individuals.

The ideal cognitive digital twin is an autonomous agent that represents the skills and knowledge of the person [2], and its digital nature facilitates the integration and collaboration with cyberspace and software systems. Although It sounds very promising, there are many hurdles to the realization of cognitive digital twins. Our proposal seeks to utilize data that users created and consumed to simulate their cognitive skills such as decision making and long-term memory.

As we know, in the past decade, information technology has integrated into our daily life more and more, and we see rapid growth of smartphone ownership globally. As the world is going to be more connected, smartphone users will create and consume more information. The management of their personal information becomes more difficult for users than ever. For example, users may forget what terms they searched on Google and their preferred results. However, Google preserves this information on its servers to use it in big data analytics.

In other words, users' digital footprint and online activities can reveal the user's preferences, which is much valuable for internet companies to risk overstepping users' privacy and regulatory rules. In this work in progress paper, we propose a system that would put back acquired intelligence to the hands of its actual owners(users).

A blockchain network as a distributed immutable ledger can facilitate the secure and persistent storage of user information while ensuring ownership of data and end users' privacy. Medical centers currently use the blockchain platforms like Medicalchain in order to store patient's health records securely [3],[4]. Similarly, for creating the personal cognitive digital twin, we need to encode all of an individual's digital activities and interactions on a blockchain-based infrastructure. In Cognitive Ledger Project, we try to mimic the human mind by collecting as much as possible data and information about users and turning them into usable information by machines. After few years, when this data covers most aspects of the individual's consciousness, similar to Ghost in the Shell world [5], the shell applications can use the cognitive blockchain to deliver the cognitive digital twin with different levels of agency and automation.

Furthermore, blockchain technology has driven the paradigm of tokenization of assets, and several companies around the world are working on blockchain projects for creating tokens that represent ownership of tangible and intangible assets [6],[7]. To put it another way, the immutability of records and the capability of issuing tokens facilitate the tokenization of knowledge and cognitive skills.

In the following sections, we introduce a cognitive architecture for representing individual intelligence. Then, we describe a blockchain ledger system that supports secure storage and representation of users' knowledge. After that, we discuss steps toward the implementation of the proposed system.

## II. PCDT 1.0 Cognitive Architecture

Before developing a cognitive blockchain ledger, we need to choose a cognitive architecture for the personal digital twin. During the last few decades, many researchers worldwide have worked hard to design architectures to represent human-like cognitive abilities in digital environments. Adams et al. [8] presented important competency areas associated with human-like general intelligence (See Fig. 1)

We needed to design a modular architecture that can be used for the creation of cognitive digital twins by third parties.

The participatory design methodology was used for defining the design requirements. The recruited participators were briefed about the cognitive digital twins, and then they were organized into two-person teams. Each teammate checked the social media profiles of their teammate and tried to play the role of the teammate cognitive digital twin by writing its logs. Then the other teammate gave feedback on the usefulness and the perceived experience from these logs. By analysis of these reports, the following design requirements were gathered:

- The architecture should store personality traits and preferences, which can lead to more humanized interactions.
- It should convey the user's online interactions, such as consumed information.
- It should mimic the decision-making pattern of the user.
- ML Trainer Engine: The programs that train personalized machine learning models. In our vision, these trained models work together to compose the cognitive digital twin of the user.

Fig 1. Artificial general intelligence competencies [8]

By considering Fig.1, design requirements, and altering CIT architecture, the PCDT 1.0 architecture (Personal Cognitive Digital Twin) was designed.

CIT architecture is developed by Chandiok and Chaturvedi [9]. They considered state of the art in cognitive architectures and created a wholly general-purpose and adaptable architecture. The other reason for using CIT architecture elements is that this architecture is developed and implemented based on the latest practical artificial intelligence technologies.

The mental layer is at the center of our proposed architecture, which is designed to be implemented on the blockchain. The information from users is captured through the information capturing module and stored in the memory pool of this layer. The collected information is raw and must get processed. The learning layer process this information through agents including:

- Knowledge Codification Agents: These autonomous agents turn the raw information into knowledge objects and personality trait badges. This process makes the collected information into a reusable form for usage by Digital Twin agent and dApps. The term of knowledge codification came from knowledge management literature and referred to transferring tacit knowledge to explicit knowledge for later use by organizations [10].
- Knowledge Improver Agents: There is an improvement phase in most knowledge management cycles that take place after initial learning [11]. Meyer and Zack, [12] used the term of the refinery for this KM activity. These agents keep the knowledge codified knowledge give weight to codified knowledge and may burn the unnecessary knowledge from the ledger.

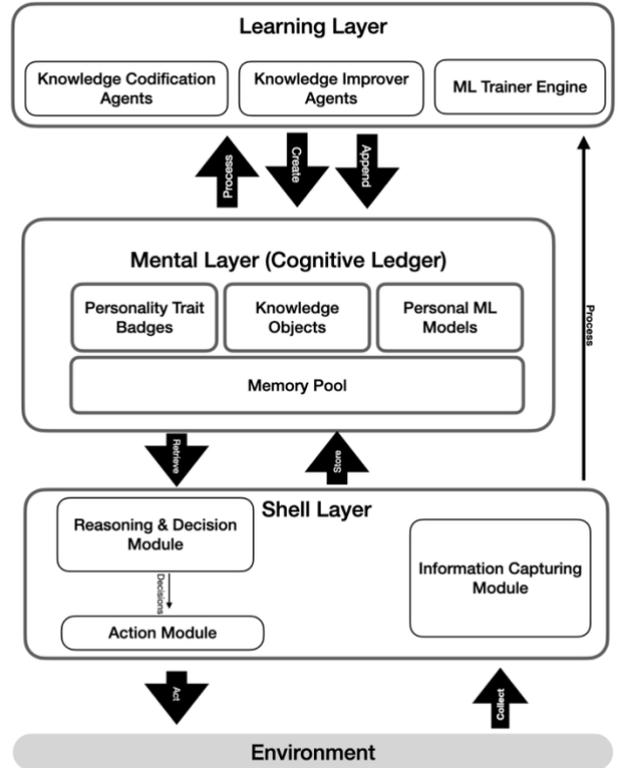

Fig 2. PCDT 1.0 Cognitive Architecture

Finally, the shell layer allows users to benefit from their virtualized knowledge on the blockchain or start new knowledge synchronization sessions. In other words, users can connect decentralized applications (dApps) to their cognitive ledger. These applications can provide different levels of AI, from no automation to fully autonomous digital twin agents.

III. COGNITIVE LEDGER ARCHITECTURE (MENTAL LAYER)

The users' activities and interactions in the digital environment, such as the terms they searched and webpages visited, are a precious source for digital cognitive eternity. And the mental layer or Cognitive Ledger is a blockchain-based infrastructure designed to store user data and intelligence.

Cognitive Ledger is a crucial foundation for the development of cognitive digital twins. From our knowledge, a critical reason that we are far from the functional cognitive digital twins is there is no infrastructure for storing all of this sensitive information about users securely. We hope that we create a basis for efforts to develop cognitive digital twins by implementing this infrastructure.

We propose that a system lifelogs and tracks the user. Then it turns the collected information into a queryable format for applications. Also, the cognitive ledger should show the knowledge assets to its owner in the same way that a cryptocurrency wallet displays assets. In addition, it should facilitate the management of the assets.

### A. Memory Pool

The cornerstone of the cognitive ledger is the memory pool. Several efforts have been conducted for secure and tamperproof log storage through private and public blockchains [13], [14], [15], [16]. The idea of immutable log storage as a service [15] seems promising for the memory pool module. But there are concerns over the size of the blockchain ledger because the blockchain data structure of blockchain requires that all nodes have a copy of the ledger. It means that each node needs to store terabytes of data after few years. However, we think we can address this issue through virtualization, so the on-chain storage approach is selected to store user activities.

### B. Personality Trait Badges

The learning agents recognize the personality traits of users from the memory pool. The personality traits are stored in the form of non-fungible tokens (NFTs). Furthermore, third party shell applications can issue personality trait badges. These tokens display the personality attributes of users, such as their personality type according to the Myers-Briggs Type Indicator [17].

### C. Knowledge Objects

In this model, knowledge objects are content resources about the user and include vocabularies and knowledge bases based on what the user knows. And they are represented as NFT Tokens. It is apparent that these files are larger than the limitation of the metadata field of blockchain records. Furthermore, some knowledge bases need to be queryable. So, storing knowledge objects requires a decentralized filesystem and a blockchain-based database system. Only the hash of the file can be saved as the metadata field of NFT tokens.

### D. Personal ML Models

Similar to knowledge objects, personal ML models are represented as NFT, too. The NFT holds the hash of the model on off-chain storage services.

Personal Models can promote the commercial aspect of the cognitive ledger. For example, the aesthetic taste of an artist can be simulated into an ML Model. This would be only one of the possible applications of mind and knowledge tokenization.

## IV. LEARNING LAYERS & SHELLS

The idealistic approach for implementing the learning layer would be integrating the knowledge mining programs and agents on blockchain nodes. It means miners would earn tokens based on the knowledge they mine for the users. The other blockchain-friendly approach would be the development of a blockchain-based decentralized marketplace of knowledge miners. It allows users to subscribe to knowledge mining services via smart contracts. Nardini et al. [18] proposed a framework for building a decentralized electronic marketplace for computing resources in 2020, so it is possible to implement this approach, but it requires in-depth evaluation and experiments.

Thus, we decided to postpone developing the knowledge mining marketplace at this early stage and apply the on-device machine learning paradigm. In contrast to centralized cloud-based systems, on-device machine learning would limit data privacy concerns, and new smart devices such as the new iPad Pro and the new ARM-based MacBook, which are capable of running on-device machine training frameworks, can reduce our concerns over the feasibility of using on-device ML.

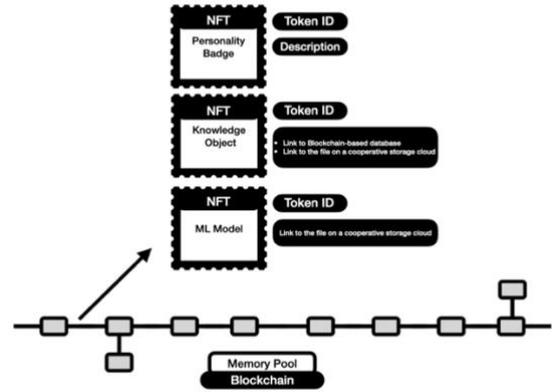

Fig 3. Cognitive Blockchain and Mental Layer

Therefore, the learning modules will be integrated into several shell modules. Also, the third-party developers can create new shells to integrate novel learning methods or use the learned knowledge to provide intelligent human-computer interactions.

The Cognitive Wallet would be the first shell that will be released, and it would be a wallet and gateway to the cognitive blockchain. It allows users to access their knowledge assets through a browser extension, which can then be used to interact with the third-party shells. It is inspired by the MetaMask wallet [19]. Furthermore, it includes some learning modules. First, it collects the links that the user visits. It also collects bookmarked URLs and the time users spent on these pages. Then with the help of Natural Language Processing pre-trained models, the web extension identifies the topics and persons mentioned on these pages. Moreover, the quiz panel recognizes the personality traits of the user.

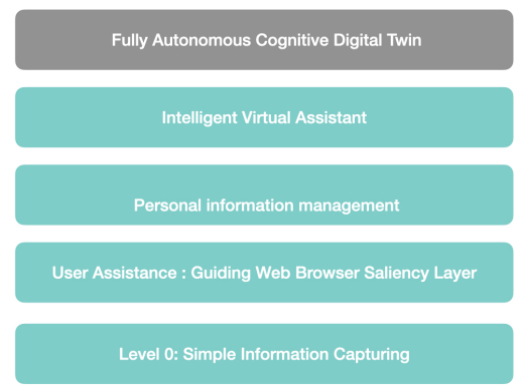

Fig 4: Different levels of Cognitive Digital Twins in Cognitive Ledger Project

Cognitive Ledger Project will work with individuals and organizations worldwide to facilitate the utilizing of the cognitive digital twins in interactions between humans and

computers. Utilizing the generated knowledge can help the user in several ways, such as:

- Intelligent virtual assistant: A virtual assistant that is made upon the memories and cognitive abilities of the user can acquire new knowledge based on the personal traits and preferences of the user. This virtual assistant will be able to make decisions and act similarly to the user in multiple ways.
- Personal information management: This shell allows users to search and explore their minds. Furthermore, It can collect information based on the user's personality traits.
- Web browsing: A browser extension queries the ML Model to add a preferred information saliency layer on websites to highlight the more significant information to the user.

## V. ON-GOING WORK AND NEXT STEPS

At this stage, we are evaluating different blockchain technologies to address the hurdles mentioned in previous sections. The blockchain network should be capable of storing memory pool records and NFTs.

The primary candidate for implementing cognitive blockchain is running a public blockchain using a custom version of Ethereum. Ethereum is widely used for minting non-fungible tokens, and there are many tools for implementing NFTs in the Ethereum blockchain. We set up a private Ethereum blockchain and compiled and deployed several smart contracts for testing the prototypes. The most critical issue about Ethereum is that it uses a consensus protocol called Proof-of-work (PoW). This protocol uses a lot of energy and computing resources, and newer platforms turned to use new consensus protocols such as Proof-of-Stake (PoS), which is designed to eliminate energy consumption [20]. Ethereum Foundation planned to update its program with a Proof-of-Stake algorithm in too, and this upgrade may make the "Ethereum miners obsolete" [21]. Once Ethereum completes this upgrade, and we have enough information about the impacts of this upgrade, we can select the platform of choice for implementing the blockchain.

For filesystem implementation, IPFS [22] and Swarm [23] are the main candidate networks. Both systems use content-based addressing and allow distributed content sharing and storage [24]. IPFS focuses on immutable data and is used as the storage layer for many blockchain projects. Swarm aims to become the storage layer of the internet too, but it is deeply integrated with Ethereum and depends on an Ethereum-based incentive mechanism. So, once the blockchain network is finalized, it would be easier to select the proper protocol for the storage layer.

Meanwhile, we gather information about the performance of blockchain networks and protocols. Selecting and developing relevant ontologies for storing the mental layer is also a part of future work.

In addition, we are constantly examining the automated machine learning (AutoML) methods for use on the learning layer. The AutoML is an inevitable piece of creating the digital cognitive twin puzzle. There are several open-source and commercial AutoML systems at this moment. Still, they provide varying levels of automation, and there is no fully autonomous AutoML system at this moment [25], and some level of user collaboration is still required.

## VI. PRIVACY CONCERNS AND RELATED WORKS

Although there is no company that actually implemented the personal digital twins, it grasps the attention of marketers for using cognitive digital twins in marketing analytics[26]. It is important to emphasize that unlike projects like Datacoup [27], we do not aim to turn the project into a personal data brokerage system. Although we allow users to fully own and manage their valuable memory and knowledge, we aimed to design an architecture that prevents unauthorized access to the cognitive ledger. Furthermore, the users' data is not connected to their contact info, and the knowledge mining process will be executed on their devices.

The critical scenario happens when the users allow closed source shell applications to use their cognitive ledgers, and the application steals this information. Reviewing the shell applications is one possible solution for addressing this scenario.

## VII. CONCLUDING REMARKS

This paper presents a modular cognitive architecture for application in the development of cognitive digital twins. The cognitive ledger is at the heart of this architecture. By lifelogging users and collecting their activities in the digital environment, it enables them to digitize their minds for usage in artificial intelligence systems. Moreover, it is evident that using blockchain-based decentralized infrastructure can ensure users on the security of their information. In other words, the cognitive ledger tries to model the user's mind into immutable records in order to be used by current and future machines. The design and development of the shells utilizing the cognitive blockchain can be the topic of future studies.

Finally, we should consider the limitations of this paper. As we noticed earlier, this paper is a work in progress, and the proposed architecture should get evaluated in more scenarios and situations. It is clear that there is a long way toward implementing a fully autonomous cognitive digital twin, but we shared this early-stage study to break the mental ceiling in this field.

### WEBSITE

For more information you can check, CognitiveLedger.org